# A Computer Vision Framework for Multi-Class Detection and Tracking in Soccer Broadcast Footage

By: Daniel Tshiani

## Abstract

Clubs with access to expensive multi-camera setups or GPS tracking systems gain a competitive advantage through detailed data, whereas lower-budget teams are often unable to collect similar information. This paper examines whether such data can instead be extracted directly from standard broadcast footage using a single-camera computer-vision pipeline. This project develops an end-to-end system that combines a YOLO object detector with the ByteTrack tracking algorithm to identify and track players, referees, goalkeepers, and the ball throughout a match. Experimental results show that the pipeline achieves high performance in detecting and tracking players and officials, with strong precision, recall, and mAP50 scores, while ball detection remains the primary challenge. Despite this limitation, our findings demonstrate that AI can extract meaningful player-level spatial information from a single broadcast camera. By reducing reliance on specialized hardware, the proposed approach enables colleges, academies, and amateur clubs to adopt scalable, data-driven analysis methods previously accessible only to professional teams, highlighting the potential for affordable computer vision–based soccer analytics.

## Introduction

Clubs that adopted data analytics early have already demonstrated its competitive value, inspiring others to follow. One of the most notable examples is Brighton & Hove Albion. Their analytics-driven recruitment strategy has produced multiple high-profile successes: Moisés Caicedo was purchased for £4 million and later sold for £115 million; Alexis Mac Allister joined for £6.9 million and was sold for £35 million; Kaoru Mitoma arrived for £2.6 million and is now valued above £50 million (Transfermarkt). These are not isolated cases but indicators of a systematic, data-informed process. Brighton's Technical Director, David Weir, highlighted this advantage when he said that the club is "fortunate to have access to all the information from every league in the world, which you'd never be able to cover on a scouting, subjective, eyes-on basis."(Kershaw) Analytics have contributed significantly to Brighton becoming one of the most profitable clubs in the Premier League during the period from 2019 to 2023 (Swiss Ramble). This context demonstrates why the integration of data and AI is transformative.

Unfortunately, data does not appear by itself. As Weir implied, it is impossible for scouts to monitor every player across every league and manually extract meaningful insights. Computer vision can offer a scalable solution. It can enable clubs to construct pipelines that automatically extract data from footage and present it in a suitable format to facilitate monitoring a larger number of players. The first stages of this pipeline would be getting artificial intelligence to detect and track players, goalkeepers, the ball, and the referee. This paper investigates that foundational step. Specifically, this study answers the question:

**Can Artificial Intelligence be used to reliably detect and track soccer players, goalkeepers, balls, and referees using a single broadcast camera?**

To address this question, we develop and evaluate an end-to-end computer vision pipeline for detecting and tracking key soccer entities from broadcast video. By applying a modern object detection and tracking framework, this study assesses the feasibility and limitations of single-camera systems for scalable soccer analytics.

The remainder of the paper is structured as follows. Section II reviews related work in prior research and industry applications. Section III outlines the original contribution of this study. Section IV describes the methodology and processing pipeline. Section V presents the results and performance of the pipeline. Section VI discussion acknowledges the limitations and outlines future research directions. Section VII concludes the paper.

## Related Work

Research on the application of computer vision to soccer has grown in recent years, although the field remains relatively underdeveloped compared to other computer vision domains. Prior work has explored player detection, tracking, and event understanding in broadcast footage, including benchmark efforts such as SoccerNet (Cioppa et al.) and early systems for player identification and tracking in sports video (Lu et al.; Thomas et al.). These studies highlight the unique challenges posed by soccer footage, including frequent occlusions, dense player interactions, and continuous camera motion.

One of the more comprehensive soccer-specific studies is by Theiner et al., who evaluated multiple detection and tracking architectures, including Faster R-CNN, FootAndBall, and CenterTrack. A key contribution of their work is the explicit consideration of broadcast-style camera movement, which reflects realistic match recording conditions. However, several of the architectures evaluated—particularly Faster R-CNN—have since been surpassed by more recent one-stage detectors such as YOLO in both inference speed and detection accuracy, limiting their suitability for real-time or large-scale analysis.

Another relevant contribution is by Markappa, who presented an extensive overview of the evolution of the YOLO family of object detection models, highlighting architectural improvements across successive generations. While this work provides useful context on detector development, it primarily focuses on comparing YOLO variants rather than evaluating their performance within soccer-specific detection and tracking pipelines.

Beyond soccer-focused research, advances in multi-object tracking (MOT) have strongly influenced modern tracking pipelines. Methods such as DeepSORT (Wojke et al.) and ByteTrack (Zhang et al.) introduced robust identity association strategies that perform well under occlusion and camera motion, conditions commonly observed in broadcast sports footage. These approaches form the foundation of many contemporary tracking systems applied across domains.

Together, these works demonstrate both the promise and limitations of existing approaches. Prior research has either emphasized realistic video conditions using older detection architectures or focused on general detector evolution without domain-specific evaluation. As a result, there remains a gap in understanding how modern, lightweight object detectors perform across different soccer-specific object classes—particularly when constrained to single-camera broadcast footage. This study seeks to address that gap by evaluating a contemporary detection and tracking pipeline tailored to soccer analysis and by examining performance differences across players, referees, goalkeepers, and the ball.

## Original Contributions

Compared to prior research, this study builds upon foundational work in soccer computer vision while advancing the methodology through the application of a modern YOLO detector. A central contribution is the empirical evaluation of detection performance across individual classes—players, referees, goalkeepers, and the ball—rather than treating all objects uniformly. Such class-specific analysis is particularly important given the limited prior work in soccer-focused computer vision and the varying difficulty of detecting different entities in broadcast footage.

To the best of our knowledge, no existing study has applied YOLOv8 to soccer footage while providing detailed, class-level performance benchmarks. By reporting metrics separately for each object type, this work establishes a more granular baseline for future research and practical applications in sports analytics, offering insights into where modern detectors perform well and where further improvements are needed.

# Methodology

## Overall Pipeline

The overall approach of this study consists of a multi-stage computer vision pipeline designed to detect and track players, referees, goalkeepers, and the ball in soccer match broadcast footage. Broadcast video serves as the input to the pipeline and is decomposed into individual frames. To reduce computational cost while preserving temporal continuity, frames are sampled using a stride of 30, corresponding to approximately one frame per second for standard broadcast footage.

Each sampled frame is processed using the YOLOv8s object detection model to localize the soccer players, goalkeepers, referees, and ball; and generate axis-aligned bounding boxes in xyxy format. Detected objects are subsequently cropped from the frame and encoded using the CLIP image encoder to obtain fixed-dimensional appearance embeddings. The CLIP transformer is used as a feature extractor to generate semantic representations of object appearances, producing embedding vectors of dimension 512.

To facilitate clustering and visualization, the high-dimensional CLIP embeddings are projected into a three-dimensional latent space using UMAP. The resulting low-dimensional representations are then clustered using K-Means to assign players to their respective teams based on their jersey colors.

Object identities are maintained across frames using ByteTrack, which associates detections over time and assigns consistent unique identifiers to tracked entities. The integration of detection, appearance-based embedding extraction, dimensionality reduction, clustering, and tracking enables not only robust object detection and tracking but also team-level segmentation for downstream analysis.

## Data

The dataset used in this study was a custom collection from Roboflow, which included annotated broadcast footage from the German Bundesliga. To increase diversity in camera angles, lighting, and skill levels, I supplemented this footage with match recordings from NCAA Division I and UPSL matches. Combining professional and semi-professional footage allowed the model to generalize more effectively across different contexts.

## Data Preprocessing

Before training, all raw videos were split into frames at a consistent frame rate. Each frame was resized and normalized according to YOLO's preprocessing standards. Additional augmentations such as horizontal flipping, rotation, brightness changes, and random cropping were applied to increase robustness to broadcast variations. All images were converted to RGB and scaled to the chosen input size of 1280×1280 pixels to ensure finer granularity when detecting small objects such as the ball.

Videos were decomposed into individual frames at their native frame rate. To reduce computational overhead while maintaining temporal coverage, frames were sampled using a stride of 30, corresponding to approximately one frame per second for standard broadcast footage.

Each sampled frame was resized and normalized according to YOLOv8 preprocessing standards. All images were converted to RGB and scaled to an input resolution of 1280 × 1280 pixels to preserve fine-grained spatial details, particularly for small objects such as the ball. No additional data augmentation was applied, as pretrained detection weights were used and all processing was performed at inference time.

## Model Architecture

The detection architecture used was the YOLOv8s model. Although training initially began with the YOLOv8x variant, hardware constraints—including GPU memory limitations—made the larger model impractical. YOLOv8s provided an optimal balance between computational efficiency and performance.

The model takes raw RGB frames through an initial convolutional layer, which converts the 3-channel image into 32 feature maps. The backbone of the network contains 127 hidden layers composed of:

- Convolutional Layers — Convolutional layers extract low- and mid-level visual features from broadcast soccer footage, such as player silhouettes, jersey textures, field markings, and ball contours. These features provide the spatial foundation for distinguishing visually similar entities (e.g., opposing teams with similar kits) and for localizing small, fast-moving objects such as the ball under varying lighting and camera conditions.
- Batch Normalization and Activation Functions — Batch normalization stabilizes intermediate feature distributions across varying lighting conditions, camera exposure changes, and broadcast quality, while nonlinear activation functions enable the network to model complex visual variations in player appearance, jersey textures, and background clutter. Together, these components improve robustness when detecting visually similar entities under dynamic match conditions.

- C2f / Cross Stage Partial (CSP) Blocks — C2f and CSP blocks enable efficient deep feature extraction by partitioning and reusing feature maps, reducing redundancy while preserving representational capacity. This design is well-suited for sports analytics because multiple objects must be detected simultaneously without incurring excessive computational cost.
- Down-sampling and Up-sampling Layers — These layers facilitate multi-scale feature learning by aggregating contextual information at coarse resolutions and recovering spatial detail at finer scales. This is critical for soccer analysis because objects of interest vary significantly in size, ranging from full-body player detections to a small, fast-moving ball.
- Spatial Pyramid Pooling Fast (SPPF) — The SPPF module captures global spatial context by pooling features at multiple receptive field sizes, enabling the model to reason across varying camera zoom levels and field coverage. This global context improves detection performance when players appear at different scales due to camera distance or perspective changes.
- Concatenation Layers — Concatenation layers fuse feature maps from multiple resolution levels, combining fine-grained spatial details (e.g., jersey edges, limb boundaries, and the ball) with higher-level semantic context (e.g., full player body structure and field layout). This multi-scale fusion improves detection in crowded match situations such as set pieces, midfield congestion, and goalmouth scrambles, where players frequently overlap or are partially occluded in broadcast soccer footage.
- Detection Head — The detection head produces bounding box coordinates, class probabilities, and confidence scores for each detected entity. In this work, these outputs serve as the basis for downstream tasks, including cropping, appearance embedding extraction, and temporal association, thereby enabling player tracking and team-level segmentation.

Together, these components form an efficient end-to-end detection pipeline optimized for soccer footage, enabling reliable object localization, temporal tracking, and downstream appearance-based analysis in real-world match settings.

### Training Procedure

To better detect small objects—specifically the ball—the input resolution was increased from the default 640×640 to 1280×1280 pixels. This higher resolution preserved finer spatial details and significantly improved ball visibility. A batch size of 6 was selected to fit CPU memory constraints and accelerate training. Training ran for 25 epochs to keep computational cost minimal while still achieving convergence. AdamW was used to optimize the detector because its decoupled weight decay allowed trained YOLOv8 representations to be preserved during fine-tuning, while still adapting effectively to the high-resolution soccer frames required for accurate player and ball localization. The object detection component was trained in this work. The YOLOv8 detector was fine-tuned on pretrained weights, whereas all other components—including CLIP, UMAP, K-Means, and ByteTrack—were used without further training.  This led to more stable training, better generalization, and less overfitting—especially when using high-resolution inputs and relatively small datasets.

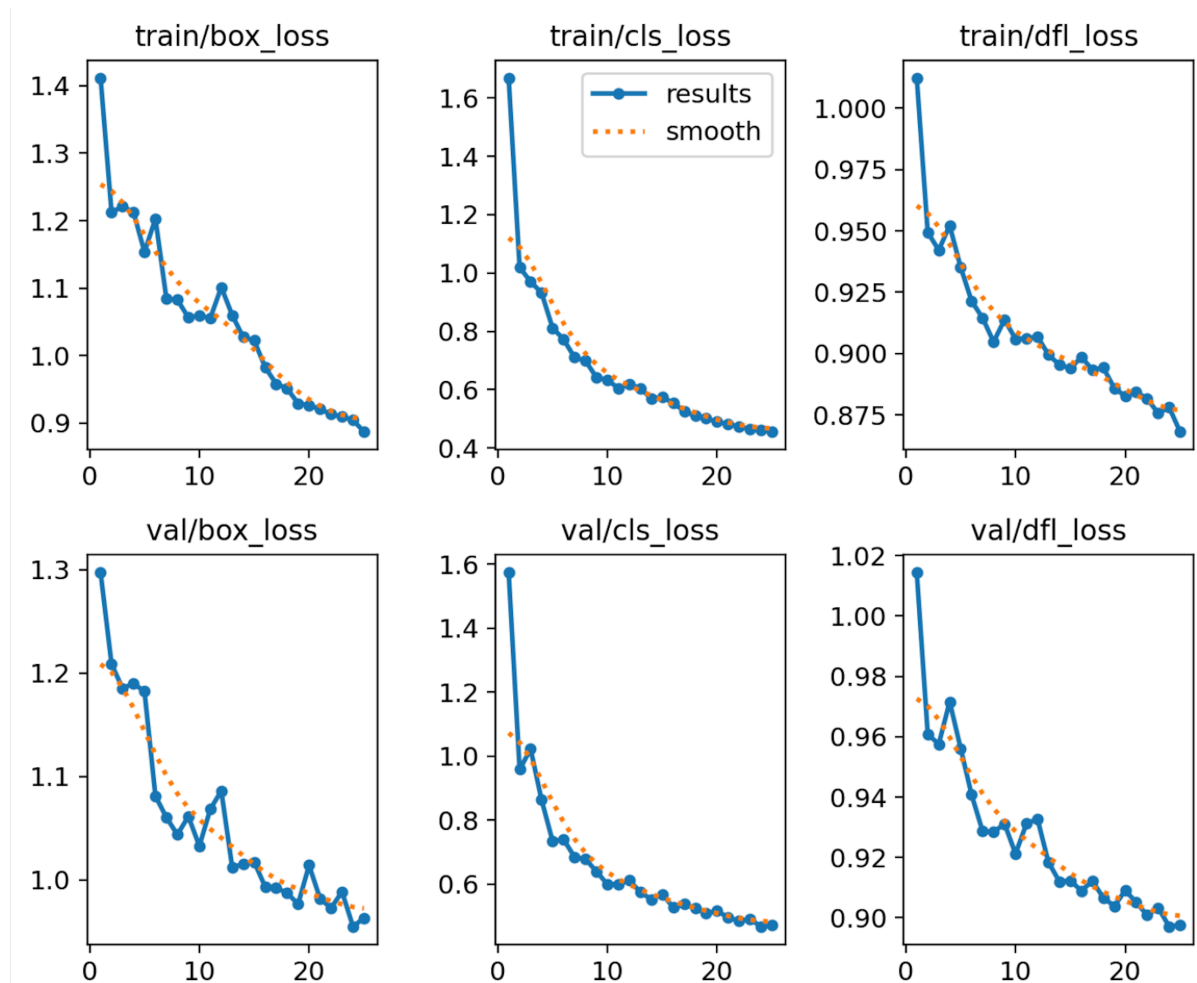

Training and validation loss curves for all three loss components decreased steadily throughout training, indicating stable convergence. Validation losses closely tracked training losses, suggesting that the model learned robust features without significant overfitting. Although both training and validation losses continued to trend downward at the end of training, the model was trained for 25 epochs due to computational and time constraints.

Moreover, three primary loss components were closely monitored during training, each tied to the unique challenges within soccer analytics:

- Box Loss — Quantifies how accurately predicted bounding boxes match ground truth, which is critical for precisely localizing players in crowded midfield scrums, goalkeepers in the penalty area, and the fast-moving ball during open play.
- Classification Loss — Measures how well the model distinguishes between visually similar entities such as players from opposing teams, referees, and goalkeepers, ensuring that team assignments and role-specific analysis remain reliable.
- Distribution Focal Loss (DFL) — Refines bounding box predictions by modeling a precise distribution of edge locations, helping the model capture small, fast-moving objects like the ball and tightly packed players near the goal.

Monitoring these losses together provided a comprehensive view of the model's learning progress, ensuring accurate detection and tracking across the dynamic and crowded scenarios typical of broadcast soccer footage.

### Evaluation Metrics

Model performance was evaluated using precision, recall, mAP@0.5, and mAP@0.5–0.95, with all metrics computed as an overall and separately for each class. Precision measured the proportion of correctly detected objects out of all predictions, reflecting the model's ability to avoid false positives — for example, mistakenly labeling a referee as a player or confusing the ball with a player's foot. Recall measured the fraction of real objects successfully detected, indicating how well the model captured fast-moving balls, overlapping players during corner kicks, or goalkeepers in the penalty area. mAP@0.5 assessed detection accuracy at an Intersection over Union (IoU) threshold of 0.5, serving as a key indicator for comparing models on soccer footage. mAP@0.5–0.95 calculated mean Average Precision across IoU thresholds from 0.5 to 0.95 in 0.05 increments, providing a nuanced view of detection quality, including challenging cases like small ball detections or tightly clustered players in the middle of the field. Together, these metrics provided a comprehensive evaluation of the model's ability to detect and distinguish all relevant soccer objects under broadcast conditions.

# Results

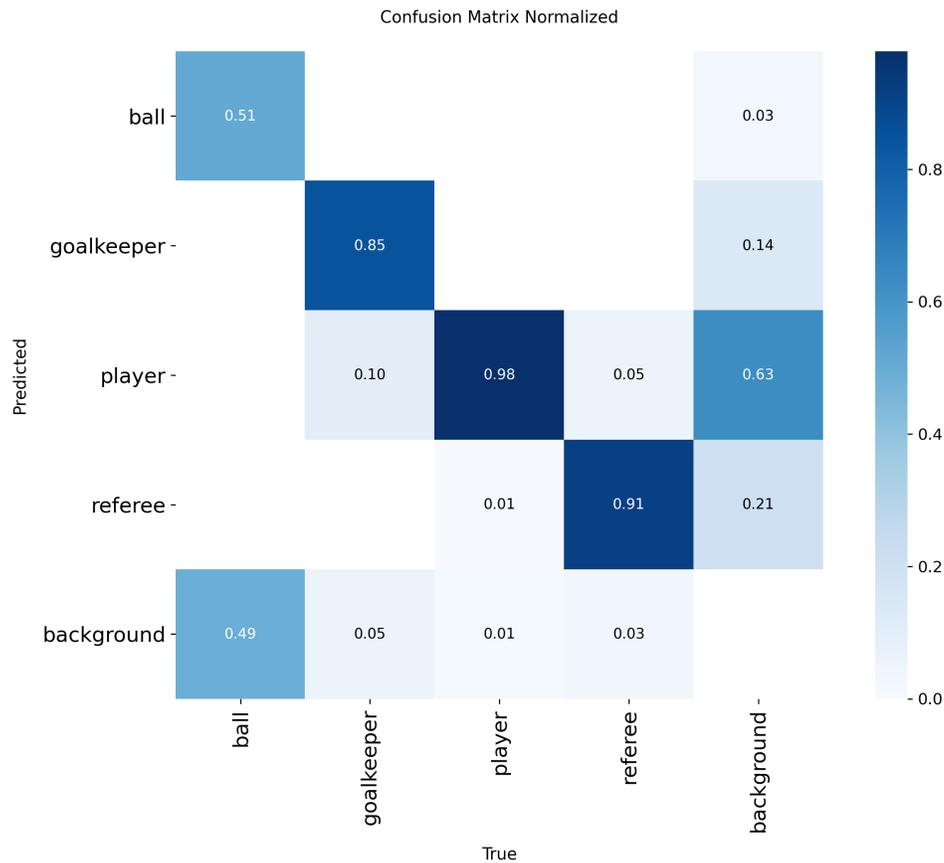

| Class | Images | Instances | Precision | Recall | mAP50 | mAP50-95 |
|---|---|---|---|---|---|---|
| All | 49 | 1174 | 0.884 | 0.828 | 0.871 | 0.583 |
| Ball | 45 | 45 | 0.914 | 0.511 | 0.616 | 0.296 |
| Goalkeeper | 38 | 39 | 0.801 | 0.897 | 0.931 | 0.659 |
| Player | 49 | 973 | 0.957 | 0.978 | 0.993 | 0.767 |
| Referee | 49 | 117 | 0.864 | 0.925 | 0.946 | 0.610 |

Across all object classes, the model achieved strong overall performance, with an mAP@0.5 of 0.871 and balanced precision (0.884) and recall (0.828). This indicates reliable detection across frames containing multiple object types simultaneously. The lower mAP@0.5–0.95 score reflects the increased difficulty of achieving very tight bounding box localization in crowded and dynamic match situations, particularly for small or partially occluded objects.

Ball detection proved to be the most challenging task, with high precision (0.914) but substantially lower recall (0.511). This suggests that, when the ball was detected, predictions were generally correct, but the model frequently failed to detect the ball. This behavior is expected given the ball's small size, rapid motion, and frequent occlusion by players' feet, especially during fast transitions and crowded play. The reduced mAP@0.5–0.95 further highlights the difficulty of precise localization for such a small, fast-moving object.

Goalkeeper detection achieved strong performance, with high recall (0.897) and an mAP@0.5 of 0.931. Despite a relatively small number of instances, the model effectively learned goalkeeper-specific visual cues such as distinct kits and consistent positioning near the goal area. The comparatively high mAP@0.5–0.95 indicates robust localization performance, even under moments where the box was crowded during corner kicks.

Player detection achieved the highest performance across all metrics, with precision (0.957), recall (0.978), and mAP@0.5 (0.993) all near saturation. This reflects the abundance of player instances in the dataset and the model's ability to robustly detect players across a wide range of poses, camera distances, and in dense formations such as midfield play and set pieces. The strong mAP@0.5–0.95 score further indicates accurate localization even under stricter IoU thresholds.

Referee detection demonstrated strong performance, with high recall (0.925) and mAP@0.5 (0.946). While referees appear less frequently than players, their distinct uniforms and consistent movement patterns enabled reliable detection. Slightly lower precision suggests occasional confusion between referees and players, particularly in crowded scenes, but overall results indicate effective identification.

## Discussion

Despite the overall strong performance, the results reveal several limitations. The most significant difficulty is in accurately detecting the ball, particularly in crowded scenes or when the camera angle obscures the object. False positives sometimes occur when cleats or bald heads resemble the ball, and false negatives arise when the ball is occluded by players' bodies. Additionally, the reliance on higher-resolution inputs (1280×1280) improves ball visibility but increases computational cost, which may limit real-time applicability on lower-resource systems. These limitations suggest potential directions for future work, including multi-frame ball reasoning, temporal smoothing, or specialized ball-focused detection modules. The result of this work demonstrates the proposed detection and tracking pipeline performs reliably; however, limitations hinder its broader applicability. In turn this can suggest directions for future work. Examples of limitations and potentially the focus of future work include long term tracking, diverse camera angles, and team classification.

ByteTrack was used to assign unique IDs to each player, but these IDs are randomly generated and do not correspond to jersey numbers. Also, they tend to be assigned a new ID when they exit and re-enter the frame. This presents a challenge for applications requiring persistent player-level statistics, such as monitoring individual performance. Ensuring players maintain consistent identities would move the system closer to real-world usability. Techniques such as appearance-based re-identification networks, sequence modeling, or temporal feature aggregation could be leveraged in future studies.

Due to the model being trained soley on broadcast-style footage, it generalizes poorly to alternative camera angles such as elevated tactical views, handheld sideline recordings, or drone perspectives. This angle sensitivity reflects both the limitations of the dataset and the difficulty of training models to handle diverse geometric distortions in sports footage. Expanding the training dataset to include footage captured from various angles would significantly improve model robustness and reduce its dependence on broadcast views.

Team classification is another limitation of the proposed pipeline. While CLIP-based appearance embeddings enabled unsupervised team clustering, performance degraded under challenging lighting conditions. In particular, players in direct sunlight were occasionally misclassified as members of the opposing team, especially when one team wore darker kits. In addition, team assignments were computed on a per-frame basis without temporal anchoring, allowing cluster identities to flip across frames. As a result, entire teams could be inconsistently labeled over time despite visually coherent player appearances. Future work could address this limitation by incorporating temporal constraints or sequence-level clustering to enforce consistent team identities across frames.

## Conclusion

This paper set out to address the core problem of detecting and tracking objects in soccer footage using only a single broadcast camera. To accomplish this, a modern YOLOv8s object detection model was employed to identify and localize individuals on the field.

The results showed that artificial intelligence can reliably detect and track players, referees, and goalkeepers, but the model struggles to track the ball. Overall, this study confirms that single-camera systems are a viable pathway to affordable, scalable soccer analytics. Nonetheless, the work provides strong evidence that AI-based pipelines can extract meaningful positional information from match footage without the costly multi-camera setups used by professional clubs.

These findings are significant for both the sports analytics research community and practical applications in soccer. For professional clubs, this work helps validate the role of computer vision in enabling automated, data-driven analysis. For colleges, academies, and amateur teams—organizations that lack access to advanced tracking technologies—this approach offers a practical, accessible path to integrating AI into scouting and performance evaluation. By lowering barriers to data collection, this work expands the potential user base for analytical tools and supports more equitable access to modern recruitment methods. Moreover, these findings establish a class-level empirical baseline for applying novel algorithms as artificial intelligence continues to advance.